\documentclass[sigconf]{acmart}
\AtBeginDocument{%
  }

\setcopyright{acmlicensed}
\copyrightyear{2018}
\acmYear{2018}
\acmDOI{XXXXXXX.XXXXXXX}
\acmConference[]{}{}{}
\acmISBN{978-1-4503-XXXX-X/2018/06}

\usepackage{enumitem}
\setlist[itemize]{label=-}

\usepackage{amsmath}
\usepackage{booktabs}
\usepackage{graphicx}
\usepackage{algorithm}
\usepackage{algpseudocode}
\usepackage{hyperref}
\usepackage{xcolor}

\title{Blue Data Intelligence Layer: Streaming Data and Agents for Multi-source Multi-modal Data-Centric Applications}
\author{
Moin Aminnaseri,
Farima Fatahi Bayat,
Nikita Bhutani, 
Jean-Flavien Bussotti,
Kevin Chan, \\
Rafael Li Chen, 
Yanlin Feng,
Jackson Hassell,
Estevam Hruschka,
Eser Kandogan,
Hannah Kim, \\
James Levine,
Seiji Maekawa,
Jalal Mahmud, 
Kushan Mitra, 
Naoki Otani, \\
Pouya Pezeshkpour,
Nima Shahbazi,
Chen Shen,
Dan Zhang
}

\email{
{moin, farima, nikita, jflavien, kevin, rafael, yanlin, jackson, 
estevam}@megagon.ai}
\email{
{eser, hannah, james, seiji, jalal, kushan, naoki, pouya, nima, chen_s, dan_z}@megagon.ai}

\renewcommand{\shortauthors}{Aminnaseri et al.}
\date{}
\begin{document}




\begin{abstract}

NL2SQL systems aim to address the growing need for natural language interaction with data. However, real-world information needs rarely map to a single SQL query because (1) users express queries iteratively across multiple utterances (2) questions often span multiple data sources beyond the closed-world assumption of a single database, and (3) queries frequently rely on commonsense or external knowledge not present in databases. Consequently, satisfying realistic data needs require integrating heterogeneous sources, modalities, and contextual data.
In this paper, we present Blue's Data Intelligence Layer (DIL) designed to support multi-source, multi-modal, and data-centric applications. Blue\cite{kandogan2025blueprint} is a compound AI system that orchestrates agents and data for enterprise settings, and DIL serves as its core data intelligence layer for agentic data processing. DIL aims to bridge the semantic gap between user intent and available information by unifying structured enterprise data, world knowledge accessible through LLMs, and  personal context obtained through interaction.
At the core of DIL is a data registry that stores metadata for diverse data sources and modalities to enable both native and natural language queries. DIL treats LLMs, the Web, and the User as source `databases', each with their own query interface, elevating them to first-class data sources. DIL relies on data planners to transform user queries into executable query plans. These plans are declarative abstractions that unify relational operators with other operators spanning multiple modalities. DIL planners support decomposition of complex requests into subqueries, retrieval from diverse sources, and finally reasoning and integration to produce final results. We demonstrate DIL through two interactive scenarios in which user queries dynamically trigger multi-source retrieval, cross-modal reasoning, and result synthesis, illustrating how compound AI systems can move beyond single-database NL2SQL.

\end{abstract}



\maketitle

\section{Introduction}

Relational database systems have long served as the backbone for managing structured data that powers modern applications \cite{plattner2009, stonebrakerpavlo2024}. Over the years, as data-driven decision-making has expanded beyond technical specialists to business analysts and domain experts, there has been growing demand for more accessible interfaces to data. This has fueled extensive research on natural language–to–SQL (NL2SQL) systems \cite{li2024, luo2025, yu2019spider}, which translate natural language questions into executable database queries.

Now, the scope of natural language data interaction is expanding to end-users with the emergence of conversational and agentic applications. In these settings, user requests rarely correspond to a single, well-formed query. Instead, they often unfold iteratively across multiple utterances, are embedded within larger task descriptions, and rely on commonsense knowledge, external facts, or personal context. These characteristics challenge the traditional closed-world assumption, in which all information required to answer a query resides within a single database schema. In practice, satisfying real-world requests often requires integrating heterogeneous data sources, modalities, and contextual data. Bridging the semantic gap between user intent and available data therefore demands mechanisms that combine structured data with external knowledge and interaction-derived context. While LLMs increasingly provide access to world knowledge, systematically integrating them with enterprise databases, organizational knowledge, and user context remains an open challenge. While novel approaches are emerging, in particular on fusing LLM knowledge with relational data, expanding to different modalities, systematically addressing optimization in multi-source settings, and tackling enterprise-scale data remain underexplored areas \cite{patel2025semanticoperatorsdeclarativemodel, alaparthi, shankar2025docetlagenticqueryrewriting, hu2025, palimpzestCIDR,russo2025abacuscostbasedoptimizersemantic, balaka, zeighami2025llmpoweredproactivedatasystems, lee2025, khoja2025weaver, shankar2026taskcascadesefficientunstructured, zhu2025relationalsemanticawaremultimodalanalytics, liu2024declarativeoptimizingaiworkloads}.


To address this gap, we present Data Intelligence Layer (DIL) as a core component of Blue. Blue is a compound AI system designed for orchestrating agents and data for enterprise applications \cite{kandogan2025blueprint}. DIL specifically aims at enabling multi-source, multi-modal, data-centric applications by treating heterogeneous information providers as first-class data sources. At its foundation is a data registry that maintains metadata across diverse sources and modalities, supporting search, discovery, and unified access through both native query languages and natural language interfaces. DIL explicitly models LLMs, the Web, and even the user as structured, queryable sources, each with its own interface and semantics. 

Beyond the registry, DIL provides a suite of data operators capable of retrieving and manipulating information across modalities, including relational, semantic, and vector representations. DIL uses planners to  construct executable data plans as directed acyclic graphs of operators. These plans decompose complex requests, retrieve data from multiple sources, and integrate results to fulfill user needs. This architecture generalizes traditional query planning to heterogeneous knowledge sources and enables compound AI systems to go beyond single-database NL2SQL toward practical, real-world scenarios.

\textbf{Contributions.} In this demonstration, we will showcase the Data Intelligence Layer of Blue through two scenarios: (1) an \textbf{apartment search} task that interactively enriches data scraped from web pages with additional data, (2) a \textbf{recipe exploration} task that begins with ingredients detected from an image of a user’s refrigerator and to retrieve recipes from heterogeneous data sources. Specifically, we will illustrate: 
(1) a \textbf{unified data source abstraction} that models LLMs, web, and users as queryable databases, (2) \textbf{data, operator and other registries} to support search and discovery, (3) \textbf{an operator framework} separating logical and physical operators across modalities, and (4) a \textbf{DAG-based multi-source multi-modal data planner} with refinement and optimization. \\
\textit{See video demonstration on https://youtu.be/Rwh6J2r4-FQ} \\
\textit{Repositories: \\
https://github.com/megagonlabs/blue \\ https://github.com/megagonlabs/blue-examples}

\section{Blue Platform: Overview}

\begin{figure}[!htb] 
  \centering
  \includegraphics[width=1.0\linewidth]{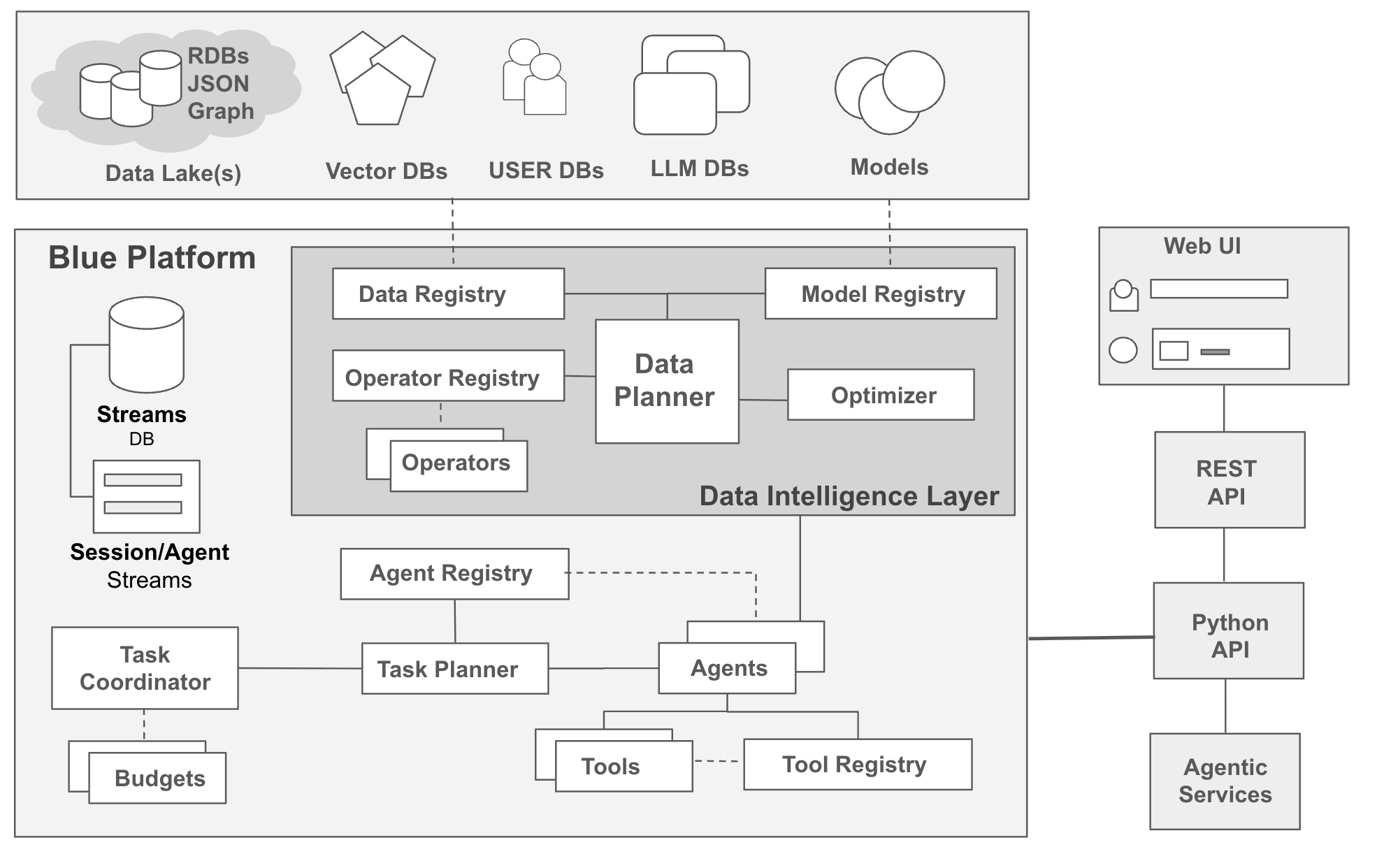}
  \caption{Blue Architecture: Registries are touch points that interface to available data, models, operators, and agents.}
  \label{fig:architecture} 
\end{figure}

Blue \cite{kandogan2025blueprint} is a prototype compound AI system \cite{chen2024llmcallsneedscaling, zaharia2024shift} for orchestrating agents and data for enterprise applications (Figure \ref{fig:architecture}). At its core, Blue organizes computation and coordination around two key concepts: \textit{streams} and \textit{sessions}. Streams are channels for data and control messages that enable agents to coordinate work and communicate results. Sessions provide contextual boundaries where agents join to accomplish tasks, share data, context, and status, and divide work across multiple agents.
Agents can register to listen to specific streams in a session, process data in the streams, and produce new streams for downstream consumption. In this way, agents are modeled as data processors, which can be implemented as LLMs, traditional algorithms, or trained predictive models.

Blue also maintains a set of registries that catalog available resources—including data, agents, models, tools, and operators. For example, the `agent registry' stores metadata for all available agents, including name, description, inputs and outputs, operational metadata such as container information, logs, execution and performance statistics, and learned metadata such as embeddings for semantic search and discovery. Similar metadata structures exist for other registries, enabling agents to locate and utilize the appropriate resources efficiently.

Within this ecosystem, task and data planners are specialized agents that leverage registries to construct executable plans. Task planners decompose high-level tasks into multiple steps, identify the agents responsible for each step, and orchestrate execution by sending control messages that direct agents to process data on specific streams. Data planners, in turn, operate on the data layer, constructing multi-source, multimodal query plans to retrieve, transform, and integrate information across diverse sources.


These components together enable Blue to coordinate complex workflows in enterprise settings, dynamically combining agent capabilities, heterogeneous data sources, and user interaction for data-driven tasks.

\section{Blue Data Intelligence Layer}

As multi-agent systems evolve to support autonomous, multi-step decision-making, data has become a critical factor in system reliability. Beyond  data access, agents need capabilities for understanding, interpreting, analyzing, modeling, planning, and communicating about data across a wide variety of sources, modalities, and systems. The Data Intelligence Layer (DIL) in Blue is designed to equip agents with these capabilities through a set of core abstractions and components, including data sources, data registry, planners, and operators. We describe each component below, using a motivating example of a multi-source, multi-modal data task.

\subsection{Motivating Example}

\begin{figure}[!htb] 
  \centering
  \includegraphics[width=1.0\linewidth]{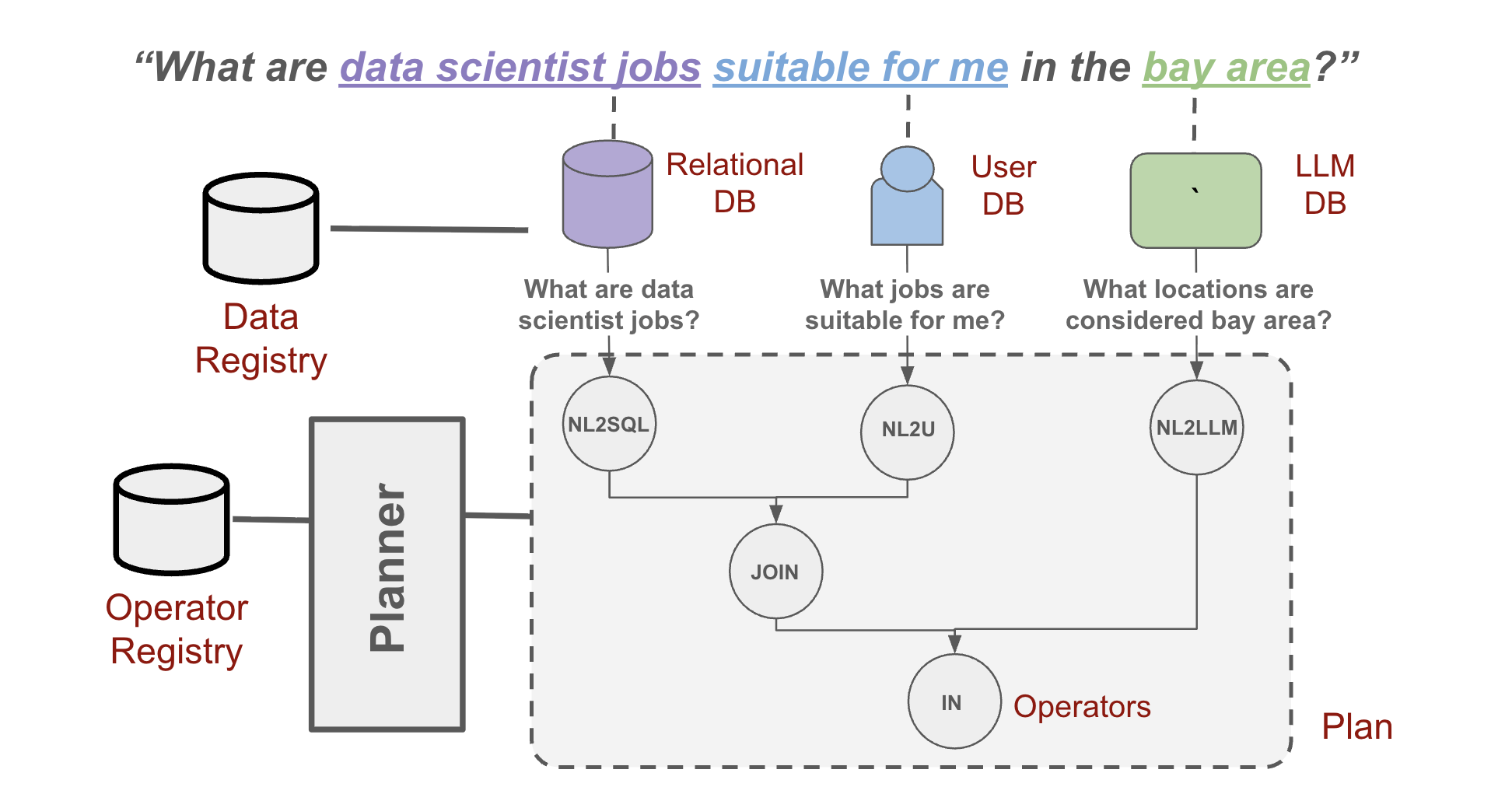}
  \caption{An example query over multiple data sources}
  \label{fig:example} 
\end{figure}

Let's consider the query: ``What are data scientist jobs suitable for me in the bay area?'' (Figure \ref{fig:example}). Answering this query inherently requires integration across multiple data sources: (1) a relational database that stores job postings, (2) a user, that provides personal preferences and requirements for the job, and potentially (3) commonsense and world knowledge to bridge the semantic gap between users query intent and available data, for example mapping `bay area' to a set of locations, or interpreting `data scientist' in terms of job title and skills.

To respond to this user query, various components in DIL work together. The \textit{data registry} catalogs available sources, including the jobs relational database, user, and LLMs. The \textit{data planner} interprets the query in terms of available sources and decomposes it into sub-queries for each source. The \textit{operators} then execute the sub-queries and integrate the results into a coherent answer. For example, the query can be decomposed as: ``what are data scientist jobs?'' that maps to jobs database (via NL2SQL), ``which locations are considered Bay Area?'' that maps to LLM (via NL2LLM) and ``what jobs are suitable for me?'' maps to the user (via NL2U). The structured outputs from each source are then integrated using additional operators, such as JOIN and IN, forming a declarative data plan that can be executed and optimized by the system.


This example illustrates how DIL enables multi-source, multi-modal queries in a unified workflow that combines structured databases, external knowledge, and user interaction. Let's now describe the role of each component in more detail.

\subsection{Data Sources}

A data source is the primary abstraction for any underlying data system. It standardizes access for agents and other components to create, update, and query data. Traditionally, these include relational database (e.g. PostgreDB), document store (e.g. MongoDB), graph database (e.g. Neo4j), and vector database (e.g. ChromaDB). To address real-world needs that go beyond traditional databases, DIL generalizes data sources to include: (1) LLMDB for commonsense and world knowledge, (2) UserDB for capturing personal data and context, (3) WebDB to capture on-demand extraction from web content. Each source is modeled as a database system with its own native query language (e.g. SQL, Cypher) as well as support for natural language queries. Furthermore, all sources expose structured outputs to allow integration across heterogeneous sources. 


\subsubsection{LLMDB}

LLMDB models LLM as a structured data source, allowing agents to query commonsense knowledge and facts using natural language. 
To facilitate accurate and performant results, LLMDB supports query rewriting, partitioning, automatic schema design, cardinality estimation, entity resolution, data/format verification, model selection, caching and distributed execution. 

\subsubsection{UserDB}

UserDB manages and retrieves structured user data through both persistent storage and interactive sessions. Initially, information ingested from sources such as personal documents is transformed into structured and vector representations, and stored for future queries. When queried, UserDB may either leverage this information or initiate an interactive session with the user to iteratively extract, clarify, and structure relevant information, subsequently storing it for future use. UserDB applies contextual judgment to assess the validity, freshness, and relevance of stored data with respect to the current query, optionally confirming with the user when uncertainty arises. 


\subsubsection{WebDB}

WebDB provides on-demand structured data extraction from web pages and documents. WebDB employs source identification, web search, scraping structured data, schema design, entity resolution, data enrichment, and distributed execution to extract structured data on demand. 


\subsection{Data Registry}

Data registry is a catalog for available data sources that supports data discovery, querying and integration. Similar to the agent registry in Blue, it contains metadata at multiple-levels of data abstractions (e.g. database, collection, entity, relation, attribute, value), independent of the underlying storage. For example, a PERSON node in a graph database is treated as equivalent to a row in a relational PEOPLE table. The registry also maintains descriptions, samples, statistics, logs, and learned representations to support agent reasoning, planning, and resolution (e.g., verification and conflict handling). Additional capabilities include data synchronization, semantic / statistical enrichment, ontology / knowledge-base alignment and value semantics discovery. Data registry functionality is exposed as tools for agents to autonomously discover, interpret, and query data across sources.

\subsection{Data Operators}

Data Operators are functions for processing heterogeneous data, including text, structured, graph, and vector data. They support data planning and optimization by enabling planners to reason over multiple operator implementations and execution strategies. Operators are organized hierarchically. Abstract (logical) operators specify the intent of an operation (e.g., extraction, query breakdown), while physical operators realize concrete implementations (e.g., dictionary-based, model-based, or LLM-based extraction). Operators expose hyper-parameters to guide execution choices such as model selection or algorithmic variants. Operators are implemented as (agentic) tools and can be invoked by any agent, including data planners, through an operator registry. Each operator is initialized with a set of properties that define execution behavior. The validity and semantics of operator invocations are captured in attributes via descriptions, explicit types and value constraints. Each operator has a standardized function signature:


\[
\mathrm{output} = \operatorname{operator}(\mathrm{input}, \mathrm{attributes}, \mathrm{properties}), where
\]
\vspace{-4mm}
\[
\mathrm{input}, \mathrm{output} \ \ \textit{typeOf} \ \ \mathrm{List}\left[\mathrm{List}\left[\mathrm{Dict}(str, Any)\right]\right]
\]
\[
\mathrm{attributes}, \mathrm{properties} \ \  \textit{typeOf} \  \ \mathrm{Dict}(str, Any)
\]

As can be seen above input and output of all operators conform to a standard and is able to represent multiple tables (e.g. list of list of dict) that supports wide range of operators (e.g. multiple table join, vector operators). The uniform function signature, where output of an operator can be fed directly to another operator, ensures interoperability across operator classes and modalities. It supports a wide taxonomy of operators, including relational, semantic, text, vector, machine learning, and custom operators. This facilitates easy design, optimization and execution of data pipelines for enterprise-scale applications. 



\subsection{Data Planning}

The DataPlanner enables agents to construct, refine, and optimize executable data-processing workflows, referred to as data plans. A data plan is a declarative specification of data workflow, represented as a directed acyclic graph (DAG) of operators that capture discovery, retrieval, transformation, and reasoning steps. Plans support abstract, alternative, and compound operators to facilitate flexible query decomposition. For example, a natural language query can be broken into subplans such as NL2LLM, NL2SQL, discovery, join, filter, or union operations.

The planning proceeds as follows. First, abstract operators (e.g. \textit{QuestionAnswer} operator) are instantiated from the Operator Registry. Next, the planner invokes a refine() function recursively to decompose abstract or compound operators into one or more alternative executable subplans (e.g. refining \textit{QuestionAnswer} into \textit{NL2SQL}, \textit{NL2LLM}, or \textit{QueryBreakdown}). This continues until all leaf nodes correspond to concrete, executable operators. The planner leverages the data, model, and operator registries to select operators and sources based on task, cost, quality, and others.


Once an initial executable plan is constructed, the \textit{optimize} phase performs operator-level and plan-level optimization. Operator-level optimization adjusts operator hyper-parameters (e.g., model choice or algorithmic variants) for efficiency or accuracy. Plan-level optimization may restructure the DAG to reduce execution cost, parallelize independent branches, or adapt to available resources. 


\section{Demonstrations}

We show two applications built with Blue that illustrate multi-source multi-modal data orchestration with agents. These applications were the top contestants of a hackathon. See Appendix \ref{appendix:survey} for analysis of a developer experience survey by participants.

\subsection{Apartment Search}


Apartment search is a complex and data-intensive task, with listings scattered across multiple websites and relevant context such as neighborhood quality often available in auxiliary sources. 

In this scenario, we demonstrate incorporating noisy, unstructured, and heterogeneous data from multiple web sources into a unified database, asking questions in natural language, and enriching the data through exploration and visualization to facilitate insights and informed decisions.
Towards this goal, the following agents are orchestrated by an interaction controller: 


\begin{itemize}
    \item DB Building from web: Scrapes apartment listings and auxiliary data from multiple websites to build or enrich a SQL database.
    \item DB Building from files: Extracts apartment or contextual information from local files to build or enrich the database.
    \item DB Transformation: Converts or restructures existing databases in the data registry.
    \item NL-to-SQL: Converts natural language questions into SQL queries for precise answers.
    \item Data Exploration: Profiles the database to highlight attribute distributions, missing values, unique entries, and potential quality issues, helping users assess coverage and reliability.
    \item Data Visualization: Generates visualizations for complex queries, such as rent trends across neighborhoods or nearby cities, to enable clearer user understanding. 
\end{itemize}


This scenario demonstrates dynamic integration of heterogeneous data sources, user interaction, and agent coordination to produce actionable insights for data-intensive tasks.

\subsection{Cooking Assistant}

Traditional recipe apps are often rigid and static, offering limited support for user-specific requirements such as available ingredients, dietary preferences, or time constraints. By contrast, LLMs can generate recipes flexibly based on user input, but their outputs may be unreliable. In this scenario, we address this trade-off by combining a verified recipe database with multi-modal user input to provide reliable, personalized recipe suggestions alongside cooking instructions and generated helper images. The workflow is as follows:

\begin{itemize}
\item Ingredient Detection: Users provide an image of their fridge. Visual recognition identifies available ingredients.
\item Candidate Recipe Retrieval: Recipes are retrieved using soft vector search (ChromaDB) based on the identified ingredients and exact filtering via relational queries (PostgreSQL)
\item Iterative Refinement: Through interactive dialogue, the system refines recipe candidates, adapting to situational constraints (e.g. missing ingredients or dietary preferences).
\item Results Presentation and Interaction: Once a recipe is chosen, the system provides step-by-step cooking instructions, answers follow-up questions, and generates helper images.
\end{itemize}

This workflow is implemented via modular sub-agents coordinated by a planner that integrates multiple data sources including structured recipe databases, recipe text, visual and textual user input, and  LLM’s internal knowledge and interactive reasoning to support complex, personalized user tasks.

\section{Conclusion}
To enable end-users effectively utilize data through natural language, agentic systems have an opportunity to bridge the semantic, structural, and modality gaps between heterogeneous data sources, common-sense knowledge, and personal context. In this paper, we presented Blue’s Data Intelligence Layer (DIL), a unified framework that treats relational databases, web data, LLMs, and users as first-class data sources, and integrates them through multiple registry-driven architecture, extensible operator hierarchy, and declarative data planning over executable DAGs. We believe this architecture represents a step toward more reliable, extensible, and semantically grounded data intelligence systems for enterprise and end-user applications alike.

\balance
\bibliographystyle{ACM-Reference-Format}
\bibliography{sample-base}

@misc{zeighami2025llmpoweredproactivedatasystems,
      title={LLM-Powered Proactive Data Systems}, 
      author={Sepanta Zeighami and Yiming Lin and Shreya Shankar and Aditya Parameswaran},
      year={2025},
      eprint={2502.13016},
      archivePrefix={arXiv},
      primaryClass={cs.DB},
      url={https://arxiv.org/abs/2502.13016}, 
}

@article{lee2025,
   title={Semantic Integrity Constraints: Declarative Guardrails for AI-Augmented Data Processing Systems},
   volume={18},
   ISSN={2150-8097},
   url={http://dx.doi.org/10.14778/3749646.3749677},
   DOI={10.14778/3749646.3749677},
   number={11},
   journal={Proceedings of the VLDB Endowment},
   publisher={Association for Computing Machinery (ACM)},
   author={Lee, Alexander W. and Chan, Justin and Fu, Michael and Kim, Nicolas and Mehta, Akshay and Raghavan, Deepti and Çetintemel, Uğur},
   year={2025},
   month=jul, pages={4073–4080} }

@misc{shankar2026taskcascadesefficientunstructured,
      title={Task Cascades for Efficient Unstructured Data Processing}, 
      author={Shreya Shankar and Sepanta Zeighami and Aditya Parameswaran},
      year={2026},
      eprint={2601.05536},
      archivePrefix={arXiv},
      primaryClass={cs.DB},
      url={https://arxiv.org/abs/2601.05536}, 
}

@misc{zhu2025relationalsemanticawaremultimodalanalytics,
      title={Beyond Relational: Semantic-Aware Multi-Modal Analytics with LLM-Native Query Optimization}, 
      author={Junhao Zhu and Lu Chen and Xiangyu Ke and Ziquan Fang and Tianyi Li and Yunjun Gao and Christian S. Jensen},
      year={2025},
      eprint={2511.19830},
      archivePrefix={arXiv},
      primaryClass={cs.DB},
      url={https://arxiv.org/abs/2511.19830}, 
}

@misc{liu2024declarativeoptimizingaiworkloads,
      title={A Declarative System for Optimizing AI Workloads}, 
      author={Chunwei Liu and Matthew Russo and Michael Cafarella and Lei Cao and Peter Baille Chen and Zui Chen and Michael Franklin and Tim Kraska and Samuel Madden and Gerardo Vitagliano},
      year={2024},
      eprint={2405.14696},
      archivePrefix={arXiv},
      primaryClass={cs.CL},
      url={https://arxiv.org/abs/2405.14696}, 
}

@article{hu2025,
author = {Hu, Chuxuan and Yang, Maxwell and Weiland, James and Lim, Yeji and Palawala, Suhas and Kang, Daniel},
title = {Drama: Unifying Data Retrieval and Analysis for Open-Domain Analytic Queries},
year = {2025},
issue_date = {December 2025},
publisher = {Association for Computing Machinery},
address = {New York, NY, USA},
volume = {3},
number = {6},
url = {https://doi.org/10.1145/3769781},
doi = {10.1145/3769781},
journal = {Proc. ACM Manag. Data},
month = dec,
articleno = {316},
numpages = {28}
}

@inproceedings{alaparthi,
author = {Alaparthi, Ashwin and Loh, Paul and Marcus, Ryan},
title = {ScaleLLM: A Technique for Scalable LLM-augmented Data Systems},
year = {2025},
isbn = {9798400715648},
publisher = {Association for Computing Machinery},
address = {New York, NY, USA},
url = {https://doi.org/10.1145/3722212.3725130},
doi = {10.1145/3722212.3725130},
booktitle = {Companion of the 2025 International Conference on Management of Data},
pages = {11–14},
numpages = {4},
location = {Berlin, Germany},
series = {SIGMOD/PODS '25}
}

@inproceedings{khoja2025weaver,
    title = "Weaver: Interweaving {SQL} and {LLM} for Table Reasoning",
    author = "Khoja, Rohit  and
      Gupta, Devanshu  and
      Fu, Yanjie  and
      Roth, Dan  and
      Gupta, Vivek",
    editor = "Christodoulopoulos, Christos  and
      Chakraborty, Tanmoy  and
      Rose, Carolyn  and
      Peng, Violet",
    booktitle = "Proceedings of the 2025 Conference on Empirical Methods in Natural Language Processing",
    month = nov,
    year = "2025",
    address = "Suzhou, China",
    publisher = "Association for Computational Linguistics",
    url = "https://aclanthology.org/2025.emnlp-main.1436/",
    doi = "10.18653/v1/2025.emnlp-main.1436",
    pages = "28282--28308",
    ISBN = "979-8-89176-332-6"
}

@article{balaka,
author = {Balaka, Muhammad Imam Luthfi and Alexander, David and Wang, Qiming and Gong, Yue and Krisnadhi, Adila and Castro Fernandez, Raul},
title = {Pneuma: Leveraging LLMs for Tabular Data Representation and Retrieval in an End-to-End System},
year = {2025},
issue_date = {June 2025},
publisher = {Association for Computing Machinery},
address = {New York, NY, USA},
volume = {3},
number = {3},
url = {https://doi.org/10.1145/3725337},
doi = {10.1145/3725337},
journal = {Proc. ACM Manag. Data},
month = jun,
articleno = {200},
numpages = {28}
}

@misc{shankar2025docetlagenticqueryrewriting,
      title={DocETL: Agentic Query Rewriting and Evaluation for Complex Document Processing}, 
      author={Shreya Shankar and Tristan Chambers and Tarak Shah and Aditya G. Parameswaran and Eugene Wu},
      year={2025},
      eprint={2410.12189},
      archivePrefix={arXiv},
      primaryClass={cs.DB},
      url={https://arxiv.org/abs/2410.12189}, 
}

@misc{patel2025semanticoperatorsdeclarativemodel,
      title={Semantic Operators: A Declarative Model for Rich, AI-based Data Processing}, 
      author={Liana Patel and Siddharth Jha and Melissa Pan and Harshit Gupta and Parth Asawa and Carlos Guestrin and Matei Zaharia},
      year={2025},
      eprint={2407.11418},
      archivePrefix={arXiv},
      primaryClass={cs.DB},
      url={https://arxiv.org/abs/2407.11418}, 
}

@inproceedings{palimpzestCIDR,
    title={Palimpzest: Optimizing AI-Powered Analytics with Declarative Query Processing},
    author={Liu, Chunwei and Russo, Matthew and Cafarella, Michael and Cao, Lei and Chen, Peter Baile and Chen, Zui and Franklin, Michael and Kraska, Tim and Madden, Samuel and Shahout, Rana and Vitagliano, Gerardo},
    booktitle = {Proceedings of the {{Conference}} on {{Innovative Database Research}} ({{CIDR}})},
    date = 2025,
}

@misc{russo2025abacuscostbasedoptimizersemantic,
      title={Abacus: A Cost-Based Optimizer for Semantic Operator Systems}, 
      author={Matthew Russo and Sivaprasad Sudhir and Gerardo Vitagliano and Chunwei Liu and Tim Kraska and Samuel Madden and Michael Cafarella},
      year={2025},
      eprint={2505.14661},
      archivePrefix={arXiv},
      primaryClass={cs.DB},
      url={https://arxiv.org/abs/2505.14661}, 
}

@misc{chen2024llmcallsneedscaling,
      title={Are More LLM Calls All You Need? Towards Scaling Laws of Compound Inference Systems}, 
      author={Lingjiao Chen and Jared Quincy Davis and Boris Hanin and Peter Bailis and Ion Stoica and Matei Zaharia and James Zou},
      year={2024},
      eprint={2403.02419},
      archivePrefix={arXiv},
      primaryClass={cs.LG},
      url={https://arxiv.org/abs/2403.02419}, 
}

@misc{zaharia2024shift,
  title={The Shift from Models to Compound AI Systems},
  author={Zaharia, Matei and Stoica, Ion and Li, Jerry and Liu, Peter and others},
  year={2024},
  month={February},
  howpublished={Berkeley Artificial Intelligence Research (BAIR) Blog},
  url={https://bair.berkeley.edu/blog/2024/02/18/compound-ai-systems/}
}

@misc{yu2019spider,
      title={Spider: A Large-Scale Human-Labeled Dataset for Complex and Cross-Domain Semantic Parsing and Text-to-SQL Task}, 
      author={Tao Yu and Rui Zhang and Kai Yang and Michihiro Yasunaga and Dongxu Wang and Zifan Li and James Ma and Irene Li and Qingning Yao and Shanelle Roman and Zilin Zhang and Dragomir Radev},
      year={2019},
      eprint={1809.08887},
      archivePrefix={arXiv},
      primaryClass={cs.CL},
      url={https://arxiv.org/abs/1809.08887}, 
}

@article{li2024,
   title={The Dawn of Natural Language to SQL: Are We Fully Ready?},
   volume={17},
   ISSN={2150-8097},
   url={http://dx.doi.org/10.14778/3681954.3682003},
   DOI={10.14778/3681954.3682003},
   number={11},
   journal={Proceedings of the VLDB Endowment},
   publisher={Association for Computing Machinery (ACM)},
   author={Li, Boyan and Luo, Yuyu and Chai, Chengliang and Li, Guoliang and Tang, Nan},
   year={2024},
   month=jul, pages={3318–3331} }

@article{luo2025,
author = {Luo, Yuyu and Li, Guoliang and Fan, Ju and Chai, Chengliang and Tang, Nan},
title = {Natural Language to SQL: State of the Art and Open Problems},
year = {2025},
issue_date = {August 2025},
publisher = {VLDB Endowment},
volume = {18},
number = {12},
issn = {2150-8097},
url = {https://doi.org/10.14778/3750601.3750696},
doi = {10.14778/3750601.3750696},
journal = {Proc. VLDB Endow.},
month = aug,
pages = {5466–5471},
numpages = {6}
}

@inproceedings{plattner2009,
author = {Plattner, Hasso},
title = {A common database approach for OLTP and OLAP using an in-memory column database},
year = {2009},
isbn = {9781605585512},
publisher = {Association for Computing Machinery},
address = {New York, NY, USA},
url = {https://doi.org/10.1145/1559845.1559846},
doi = {10.1145/1559845.1559846},
booktitle = {Proceedings of the 2009 ACM SIGMOD International Conference on Management of Data},
pages = {1–2},
numpages = {2},
location = {Providence, Rhode Island, USA},
series = {SIGMOD '09}
}

@article{stonebrakerpavlo2024,
author = {Stonebraker, Michael and Pavlo, Andrew},
title = {What Goes Around Comes Around... And Around...},
year = {2024},
issue_date = {June 2024},
publisher = {Association for Computing Machinery},
address = {New York, NY, USA},
volume = {53},
number = {2},
issn = {0163-5808},
url = {https://doi.org/10.1145/3685980.3685984},
doi = {10.1145/3685980.3685984},
journal = {SIGMOD Rec.},
month = jul,
pages = {21–37},
numpages = {17}
}

@INPROCEEDINGS{kandogan2025blueprint,
  author={Kandogan, Eser and Bhutani, Nikita and Zhang, Dan and Chen, Rafael Li and Gurajada, Sairam and Hruschka, Estevam},
  booktitle={2025 IEEE 41st International Conference on Data Engineering Workshops (ICDEW)}, 
  title={Orchestrating Agents and Data for Enterprise: A Blueprint Architecture for Compound AI}, 
  year={2025},
  volume={},
  number={},
  pages={18-27},
  doi={10.1109/ICDEW67478.2025.00007}}

\appendix

\section{Hackathon Developer Experience Survey}
\label{appendix:survey}

\subsection{Study}

We conducted a  study to investigate the developer experience of building AI-driven applications using the Blue platform after a week-long hackathon. The goal of the study is to understand how developers with different backgrounds interact with Blue’s abstractions, tools, and workflows while developing prototype systems. Participants who took part in the hackathon completed a structured survey capturing their professional background, prior familiarity with relevant technologies (such as text-to-query systems, retrieval-augmented generation, and multi-agent systems), and their previous exposure to the Blue platform. 

The questionnaire aims to evaluate several aspects of the developer experience, including on-boarding and learnability, the usability of Blue’s data abstractions (such as the data registry, data sources, and data planning components), and the broader development workflow. Participants assess how intuitive these concepts were, how easily they could discover and use available data resources, and how effectively they could integrate query results into agents. Additional questions capture friction points related to documentation, debugging, APIs, performance, and deployment. The survey concludes with open-ended reflections on the platform’s strengths, weaknesses, and suggested improvements. Together, these responses provide both quantitative and qualitative insights into the usability and effectiveness of Blue as a development environment for building data-centric AI applications.

\subsection{Results and Analysis}

\subsubsection{Background}
A total of 12 participants completed the developer experience survey. The majority were research-oriented practitioners, including 7 researchers (58\%) and 3 research engineers (25\%), with the remaining 2 participants (17\%) in product roles. Most participants had moderate professional experience, with 8 participants (67\%) reporting 3–5 years of experience, while 2 (17\%) had more than 10 years, 1 (8\%) had 6–10 years, and 1 (8\%) had 0–2 years of experience. Prior technical exposure was high: 11 participants (92\%) had experience with LLM-based agents, 8 (67\%) with RAG systems, 7 (58\%) with databases, 7 (58\%) with multi-agent systems, and 6 (50\%) with text-to-query systems such as NL2SQL. Despite this strong technical background, prior familiarity with the Blue platform was relatively low, with a mean self-reported familiarity of 2.83 out of 5 (median = 3), indicating that most participants entered the hackathon with limited direct experience using the platform.

\subsubsection{On-boarding and Learnability}

Among the participants perceptions of learnability were mixed. While half of the respondents (6/12) agreed that the overall architecture was easy to understand (scores of 4), the remaining participants reported neutral or negative perceptions (scores 2–3), and several required multiple days to become fully productive. Qualitative responses indicate that the streaming-based communication model and agent input/output handling were the most commonly cited sources of confusion and may pose a notable barrier to rapid onboarding.
Across the 12 participants, self-reported familiarity increased from an average of 2.83 before the hackathon to 3.75 after (1: "not at all, 5: "very familiar"). Eight participants reported improved familiarity (gains of 1–3 points), three (with score 4) reported no change.

\subsubsection{Data Concepts}

\begin{figure}[!htb] 
  \centering
  \includegraphics[width=1.0\linewidth]{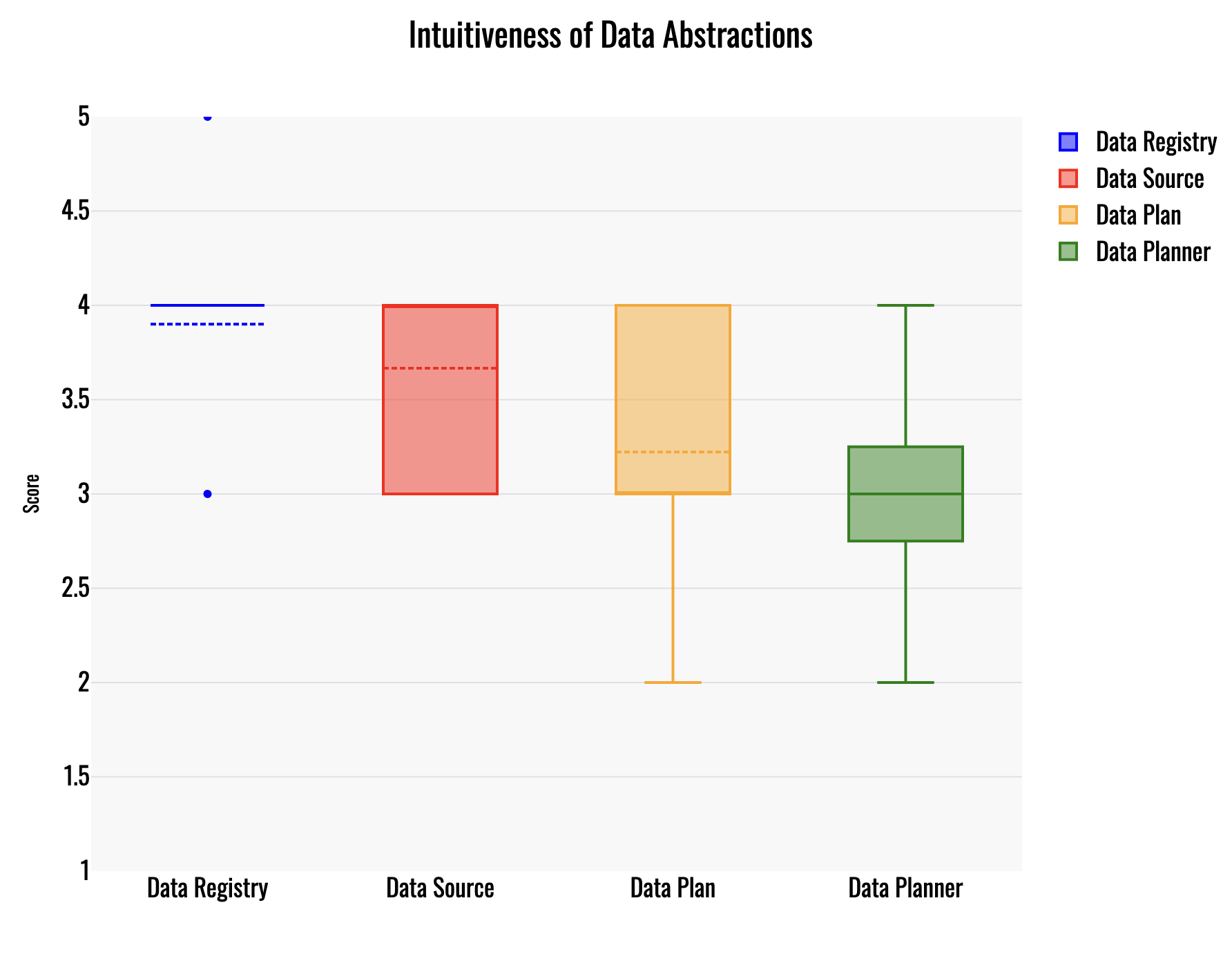}
  \caption{Intuitiveness of data abstractions}
  \label{fig:data_chart} 
\end{figure}

Blue’s core data abstractions were generally perceived as moderately intuitive (\ref{fig:data_chart}), where complex data abstractions (e.g. data plan, planner) were somewhat less intuitive than basic registry and source abstractions. In regards to data related tasks, participants reported strong ease in discovering data sources (mean = 4.5) and understanding registry metadata (mean = 4.0) while operational tasks such as connecting/querying sources (3.8) and processing results in agents (mean = 3.6) were moderately easy. Only half of the subjects reported frictions mainly on  initial connection/setup, registering datasets, and table inspection.

\subsubsection{Developer Experience}

The development experience was generally rated as moderate , with a mean score of approximately 3.1 on a 1–5 scale (1: "very poor", 5: "excellent"). Debugging was the most challenging aspect, with an average score of 2.3, reflecting frequent issues such as poor error messages, complex setups, and difficulties tracing errors across multi-agent or microservices. User interface design was less problematic, receiving an average score of 3.5, although participants noted the lack of templates or documentation as a minor friction point. The most commonly reported sources of development friction were complex setup/installation (8/12 participants), poor error messages/debugging support (7/12), and confusing or missing documentation (6/12). In regards to deployment, experience was generally rated positively, with a mean score of 3.83, though a few participants reported significant challenges, including difficulties with docker configuration and unclear deployment status.

\subsubsection{Open Feedback}

Participants' overall experience with Blue varied but highlighted strong system flexibility and modularity. Many participants praised aspects such as “flexibility of the system architecture,” “interface is so intuitive, also, adding agents and connecting them was easy,” and “modular architecture covering agents/tools/data/... you can plug in your components easily,” while common challenges included documentation and the learning curve, with comments like “lack of documentation,” “complexity of installation, deployment, and development,” and “learning curve is a bit steep.” Several participants suggested prioritizing better onboarding and clearer guides, quickstart tips. 

\subsection{Findings and Discussion}

In summary, while Blue’s architecture and flexibility were highly appreciated, practical usability improvements—especially around setup, documentation, and easing the learning curve—were consistently recommended. Participants highlighted that the platform’s modularity and data abstractions enabled experimentation and rapid prototyping, yet several friction points limited productivity, particularly in debugging, error tracing, and orchestrating multi-agent workflows. 

The hackathon experience further revealed the inherent challenges of multi-agentic development in general on distributed platforms. Participants reported difficulties in reasoning about the control flow and data flow across interacting agents, especially when multiple agents executed in parallel and shared common data resources. Debugging emerged as a core challenge, since errors often propagated asynchronously or were only visible in downstream agents, complicating the identification of root causes. Coordination of agents, monitoring execution states, and ensuring data consistency were additional pain points that affected both productivity and confidence in experimental results. These observations suggest that future iterations of Blue—or similar multi-agent development environments—could benefit from enhanced tooling for visualizing agent interactions, logging and tracing multi-agent execution, and supporting deterministic replay of workflows. Other potential directions include improved error messaging for asynchronous operations, and integration of simulation or sandbox modes to test complex agent orchestration before deployment. Addressing these challenges can make multi-agentic AI development more approachable and reduce friction for both novice and experienced developers.

\end{document}